\documentclass[runningheads]{llncs}
\usepackage{graphicx}
\usepackage{graphicx}
\usepackage{amsmath}
\usepackage{amssymb}
\usepackage{booktabs}

\usepackage{multirow}
\usepackage[pagebackref,breaklinks,colorlinks]{hyperref}
\newcommand{\tabincell}[2]{\begin{tabular}{@{}#1@{}}#2\end{tabular}}

\usepackage[capitalize]{cleveref}
\crefname{section}{Sec.}{Secs.}
\Crefname{section}{Section}{Sections}
\Crefname{table}{Table}{Tables}
\crefname{table}{Tab.}{Tabs.}

\begin{document}
\title{Visual Information Extraction in the Wild: Practical Dataset and End-to-end Solution}
\titlerunning{Visual Information Extraction in the Wild}
%

%

%
\author{Jianfeng Kuang\inst{1*}\and
Wei Hua\inst{1*}\and
Dingkang Liang\inst{1}\and 
Mingkun Yang\inst{1}\and
Deqiang Jiang\inst{2}\and
Bo Ren\inst{2}\and
Xiang Bai$^{1\dag}$}
\authorrunning{J. Kuang, W. Hua et al.}
%
\institute{Huazhong University of Science and Technology \and
Tencent YouTu Lab \\
\email{\{kuangjf, whua\_hust, dkliang, \\ yangmingkun, xbai\}@hust.edu.cn \\ 
 \{dqiangjiang,timren\}@tencent.com} \\
}
\protect \renewcommand{\thefootnote}{\fnsymbol{footnote}}
\footnotetext[1]{Equal contribution. $^{\dag}$Corresponding author.}
\footnotetext[0]{Work done when Wei Hua was an intern at Tencent.}

\maketitle              
\begin{abstract}
Visual information extraction (VIE), which aims to simultaneously perform OCR and information extraction in a unified framework, has drawn increasing attention due to its essential role in various applications like understanding receipts, goods, and traffic signs. However, as existing benchmark datasets for VIE mainly consist of document images without the adequate diversity of layout structures, background disturbs, and entity categories, they cannot fully reveal the challenges of real-world applications. In this paper, we propose a large-scale dataset consisting of camera images for VIE, which contains not only the larger variance of layout, backgrounds, and fonts but also much more types of entities. Besides, we propose a novel framework for end-to-end VIE that combines the stages of OCR and information extraction in an end-to-end learning fashion. Different from the previous end-to-end approaches that directly adopt OCR features as the input of an information extraction module, we propose to use contrastive learning to narrow the semantic gap caused by the difference between the tasks of OCR and information extraction. We evaluate the existing end-to-end methods for VIE on the proposed dataset and observe that the performance of these methods has a distinguishable drop from SROIE (a widely used English dataset) to our proposed dataset due to the larger variance of layout and entities. These results demonstrate our dataset is more practical for promoting advanced VIE algorithms. In addition, experiments demonstrate that the proposed VIE method consistently achieves the obvious performance gains on the proposed and SROIE datasets. The code and dataset will be available at \url{https://github.com/jfkuang/CFAM}.
\end{abstract}
\begin{figure}[t]
	\begin{center}
		\includegraphics[width=0.96\linewidth]{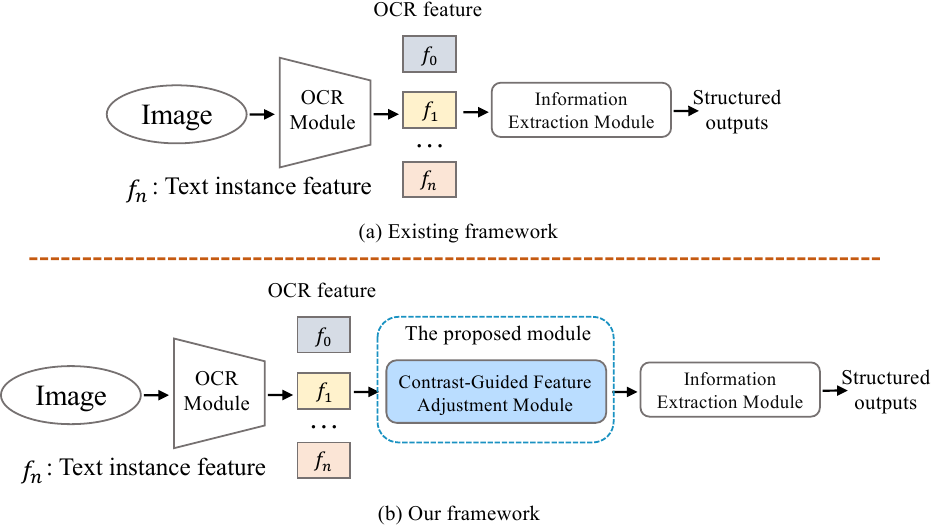}
	\end{center}
	\caption{The overview of the existing framework and our framework. The existing framework directly adopts OCR features as the input of an information module, while our framework additionally proposes a contrast-guided feature adjustment module to the existing framework.}
	\label{fig:framework}
\end{figure}

\begin{figure*}[t]
	\begin{center}
		\includegraphics[width=0.94\linewidth]{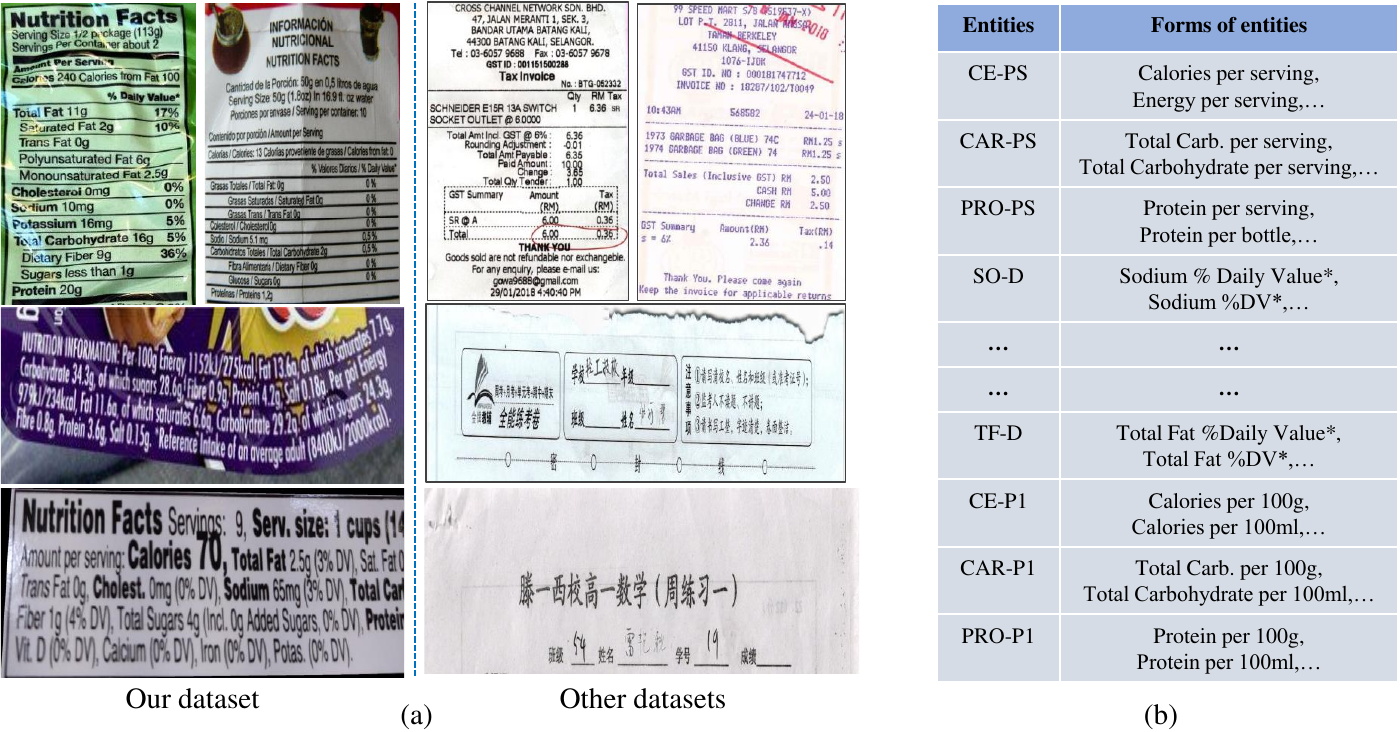}
	\end{center}
	\caption{ (a) exhibits a few typical examples from our dataset and others. (b) shows some typical entities and their different forms in our dataset. Note that the actual entity name is very long, so we use the abbreviation of the entity name. More details of our POIE dataset can be found at \url{https://github.com/jfkuang/CFAM}}
	\label{fig:datasets}
\end{figure*}

\begin{table*}[t]
\small
\setlength{\tabcolsep}{0.3mm}
\centering
\caption{Statistics of the representative VIE datasets and our proposed dataset. 
}
\label{tab:statistics of VIE dataset}
\resizebox{0.95\linewidth}{!}{
\begin{tabular}{ c|c|c|c|c|c|c|c|c }
 \toprule
 {\multirow{2}{*}{Dataset}} & {\multirow{2}{*}{Year}}& \multicolumn{3}{c|}{Number of images}& \multirow{2}{*}{\tabincell{c}{Number \\of entities}} & \multirow{2}{*}{\tabincell{c}{Number \\of instance}} & {\multirow{2}{*}{Language}}& {\multirow{2}{*}{Scene Type}}\\
\cmidrule{3-5}
 & & Train& Test & Total & & & \\
\midrule
SROIE~\cite{huang2019icdar2019} &ICDAR2019&626&347&973&4&52,316&EN&Receipt\\
EPHOIE~\cite{wang2021towards} &AAAI2021&1,183&311&1,494&10&15,771&CN&Examination Paper\\
POIE (\textbf{Ours})&-&\textbf{2,250}&\textbf{750}&\textbf{3,000}&\textbf{21}&\textbf{111,155}&EN&Product\\
\bottomrule
\end{tabular}}
\end{table*}

\section{Introduction}
Visual Information Extraction (VIE), automatically extracting structured information from visually-rich document images, is an essential step towards document intelligence~\cite{zhu2022towards,wang2022mmlayout,jiang2022revisiting}. It involves extracting values of entities from images and reasoning about their relations, which is a challenging cross-domain problem that requires both visual and textual understanding, such as table comprehension and document analysis~\cite{kahraman2010development,wong1982document,zeng2023beyond,wang2022knowledge}. This task becomes more difficult in real-world scenarios where diverse layouts, noisy backgrounds, and large variances of entities exhibit in the wild.

Datasets play an essential role in data-driven problems. In the document VIE area, a number of datasets~\cite{huang2019icdar2019,park2019cord,wang2021towards,cao2022query}, have been proposed. Specifically, SROIE~\cite{huang2019icdar2019} is the most widely used dataset, in which
the images are scanned receipts printed in English. 
EPHOIE~\cite{wang2021towards} is collected for Chinese document VIE, which is thus unsuitable for assessing previous methods for English VIE. These datasets are composed of only document images with moderate scale.
For more general scenarios, like VIE in the wild, the captured images have larger variances in layouts, backgrounds, and fonts, as well as various entities. Consequently, the available datasets cannot fully reveal the challenges of real-world applications.

To better study VIE in the wild, we collect a large-scale challenging dataset called \textbf{P}roducts for \textbf{O}CR and \textbf{I}nformation \textbf{E}xtraction (POIE), which consists of camera images from Nutrition Facts label of products in English. Compared with existing VIE datasets~\cite{huang2019icdar2019,wang2021towards}, POIE has several distinct characteristics. First, it is the largest VIE dataset with high-quality annotations, where 3,000 images with 111,155 text instances are collected. As indicated in Table~\ref{tab:statistics of VIE dataset}, it is over $2\times$ larger than the previous largest public VIE dataset. Second, POIE is particularly challenging, where the images are in diverse layouts and distorted with various folds, bends, deformations as well as perspectives. We show some typical examples in Figure~\ref{fig:datasets} (a). Third, POIE has up to 21 kinds of entities, some of which come in numerous forms. For example, in Figure~\ref{fig:datasets} (b), several forms of some entities are given. It is quite common in the real-world applications that each entity presents in various forms, demonstrating VIE in the wild is more difficult.

We observe that previous end-to-end approaches~\cite{zhang2020trie,wang2021towards} achieve lower performance on POIE, especially on the information extraction task, due to complex layouts and variable entities. Besides, these methods directly adopt OCR features as the input of the following information extraction module. We argue that the reason is there is a severe semantic gap between the tasks of OCR and information extraction while directly feeding OCR features into the following information extraction module.
Recently, we have witnessed the rise of contrastive learning in computer vision, which has received extensive attention in various visual tasks~\cite{khosla2020supervised,radford2021learning,aberdam2021sequence,huang2021model,yu2023turning} while has rarely been noticed in VIE tasks. In this paper, we propose a novel end-to-end framework for VIE. We adopt contrastive learning to effectively establish the connections between the tasks of OCR and information extraction. Specifically, the key component of our framework is a plug-and-play Contrast-guided Feature Adjustment Module (CFAM), as shown in Figure~\ref{fig:framework}. In CFAM, we design the feature representation for OCR and information extraction (i.e., instance features and entity features for OCR and information extraction). As a result, CFAM constructs a similarity matrix reflecting the relations between entity features and instance features to adjust the instance features more appropriately for the following information extraction task. 

In summary, the main contributions of this paper are two-fold: 1) we propose a large-scale dataset consisting of camera images with variable layouts, backgrounds, fonts, and much more types of entities for VIE in the wild. 2) we design a novel end-to-end framework with a plug-and-play CFAM for VIE, which adopts contrastive learning to narrow the semantic gap caused by the difference between the tasks of OCR and information extraction. 

\section{Related Works}
\subsection{VIE Datasets}
Currently, a few datasets~\cite{huang2019icdar2019,sun2021spatial,park2019cord,wang2021towards,cao2022query} are proposed for VIE on document images. Specifically, SROIE~\cite{huang2019icdar2019} is the most widely used dataset, which significantly promotes the development of this field. The images of SROIE are scanned receipts in printed English. Each image is associated with complete OCR annotations and the values of four key text fields.
Besides, EPHOIE~\cite{wang2021towards} is collected for Chinese document VIE, where the images have complex backgrounds and diverse text styles. The details of these datasets mentioned above are shown in Table \ref{tab:statistics of VIE dataset}. We can observe that these existing datasets are composed of only document images with moderate scale. To better explore VIE in the wild, a large-scale dataset with more entities and larger variances in layouts, backgrounds, and fronts is urgently required.
\subsection{VIE Methods}
According to the pipeline of VIE, existing approaches~\cite{xu2020layoutlm,yu2021pick,cao2022query,tangmatchvie,li2022table} can be divided into two categories: methods in two-stage and end-to-end paradigms. Most works with a two-stage pipeline focus on the second stage for information extraction. In these methods, the OCR results are first obtained via an individual OCR extractor. In Post-OCR parsing~\cite{hwang2019post}, the coordinates of text bounding boxes are applied during the second stage. To better model the layout structure and visual cues of documents, LayoutLM~\cite{xu2020layoutlm} employed a pre-training strategy inspired by BERT. GraphIE~\cite{qian2019graphie} and PICK~\cite{yu2021pick} constructed a graph according to the OCR results and applied Graph Neural Networks (GNNs) to extract the global representation for further improvement. In CharGrid~\cite{katti2018chargrid}, CNN is used to integrate semantic clues and the layout information. MatchVIE~\cite{tangmatchvie} noticed the importance of modeling the relationships between entities and text instances. However, it required additional annotations of all key-value pairs. All of these methods concentrated on the context modeling between OCR results in the second stage but ignored the accumulative errors from the preceding OCR module. 

Recently, an increasing number of VIE methods have been proposed in an end-to-end fashion. EATEN~\cite{guo2019eaten} first generated feature maps from input images and attached several entity-aware decoders to predict all entities. However, it can only handle documents with a fixed layout. TRIE~\cite{zhang2020trie} was an end-to-end trainable framework to solve the VIE task, which focused more on the performance of entity extraction. VIES~\cite{wang2021towards} improved each part of VIE, like text detection, recognition, and information extraction, but incurred obvious costs. Donut~\cite{kim2022ocr} can directly extract information from the input images without text spotting, while it needed massive data for pre-training. All of the above methods directly took OCR features as the input of the following information extraction module. Different from the previous end-to-end approaches, we propose a novel framework in an end-to-end manner to establish the connections between the tasks of OCR and information extraction.

\subsection{Contrastive Learning}
Contrastive learning~\cite{khosla2020supervised}, a typical way for visual representation learning, has attracted lots of attention in many fields and obtained great progress~\cite{radford2021learning,you2020graph,aberdam2021sequence}. Contrastive learning allows samples of positive pairs to lie close together in the latent space, while samples belonging to negative pairs are repelled in the latent space. CLIP~\cite{radford2021learning} is a representative work for vision-language pre-training via contrastive learning on image-text pairs. Since then, plenty of multi-modal contrastive learning methods have been proposed~\cite{patashnik2021styleclip,rao2022denseclip,zhou2022learning,li2021unimo,wang2021scene,yang2022reading}. However, to the best of our knowledge, contrastive learning has not been thoroughly studied in the VIE field. In this paper, we propose to use contrastive learning to supervise the construction of relations between the tasks of OCR and information extraction. Different from the above approaches, where the image features and text features are taken as inputs, only the features of text instances are fed into the information extraction module in the VIE task. Therefore, how to represent OCR 
features as well as entity features, and construct the relations between them to narrow the semantic gap should be of great importance for end-to-end VIE.

\section{POIE Dataset}
\subsection{Data Collection}
\textbf{P}roducts for \textbf{O}CR and \textbf{I}nformation \textbf{E}xtraction (POIE) dataset derives from camera images of various products in the real world. The images are carefully selected and manually annotated.
Our labeling team consists of 8 experienced labelers. We first crop the nutrition tables from product images and adopt multiple commercial OCR engines (Azure and Baidu OCR) for pre-labeling. Then we use LabelMe~\footnote{https://github.com/wkentaro/labelme} to manually check the annotation of the location as well as transcription of every text box, and the values of entities for all the text in the images and repaired the OCR errors found. After discarding low-quality and blurred images, we obtain 3,000 images with 111,155 text instances.
\subsection{Data Characteristics}
To the best of our knowledge, POIE is the largest dataset with both OCR and VIE annotations for the end-to-end VIE in the wild. The images in POIE contain Nutrition Facts labels from various commodities in the real world, which have larger variances in layout, severe distortion, noisy backgrounds, and more types of entities than existing datasets. The comparison of dataset statistics is shown in Table~\ref{tab:statistics of VIE dataset}. Existing datasets mainly consist of document images with insufficient diversities of layout, background disturbs, and entity categories. Therefore, they cannot fully illustrate the challenges of some practical applications, like VIE on the Nutrition Facts label. Compared with these datasets, POIE contains images with variable appearances and styles (such as structured, semi-structured, and unstructured styles), complex layouts, and noisy backgrounds distorted by folds, bends, deformations, and perspectives (typical examples are shown in Figure~\ref{fig:datasets} (a)). Moreover, the types of entities in POIE reach 21, and a few entities have different forms (some typical entities with various forms are shown in Figure~\ref{fig:datasets} (b)), which is very common and pretty challenging for VIE in the wild. Besides there are often multiple words in each entity, which appears zero or once in every image. These properties mentioned above can help enhance the robustness and generalization of VIE models to better cope with more challenging applications.

\subsection{Data Split and Evaluation Protocol}

POIE is divided into training and testing sets, with 2,250 and 750 images, respectively. We use the performance of detection (DET), recognition (REC), and information extraction (IE) in the end-to-end pipeline to evaluate all methods on the proposed POIE. Following the settings of SROIE~\cite{huang2019icdar2019} and EPHOIE~\cite{wang2021towards}, F1-Score is applied as the evaluation metric for the three tasks.

\begin{figure*}[t]
	\begin{center}
		\includegraphics[width=0.96\linewidth]{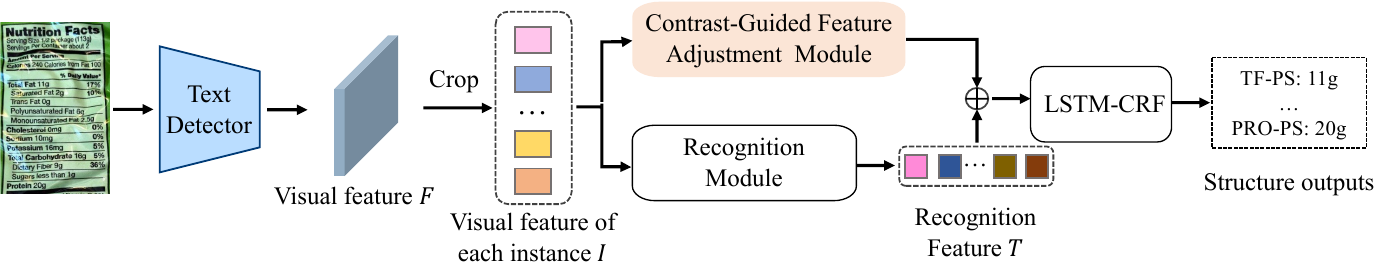}
	\end{center}
	\caption{The overall framework of our method. It consists of a text detector, a contrast-guided feature adjustment module, a recognition module, and an information extraction module.}
	\label{fig:Pipeline_method}
\end{figure*}

\section{Our Method}
The overall pipeline of our method is illustrated in Figure~\ref{fig:Pipeline_method}.
Different from other methods, we introduce contrastive learning into the end-to-end trainable framework for effectively bridging the modules of OCR and information extraction. Specifically, it is composed of a text detector, a Contrast-guided Feature Adjustment Module (CFAM), a recognizer module, and an information extraction module. Given an input image,
\begin{itemize}
    \item The CNN-based text detector not only localizes all text instances but also yields visual feature maps $F$ for subsequent modules. The text detector detects at the line level.
    \item With RoIAlign, the visual features of each instance $I$ are obtained from $F$. Then, $I$ is simultaneously fed into CFAM and a recognition module to generate a similarity matrix $S$ and the recognition features $T$. In the recognition module which is two-layers transformer decoder, we use the intermediate features in transformer as recognition features $T$ and obtain the recognition results of all text. In the CFAM, $I$ is first encoded to $E$. Meanwhile, we use a serial of learnable vectors as the entity features $L'$. Subsequently, the pair-wise similarity between  $L'$ and $E$ is calculated.
    \item Finally, the recognition features $T$, the encoded features $E$, and similarity matrix $S$ are added, and the results are fed into the following information extraction module (LSTM-CRF) to predict the structural outputs.
\end{itemize}

\subsection{Contrast-guided Feature Adjustment Module}
\begin{figure}[t]
	\begin{center}
		\includegraphics[width=0.96\linewidth]{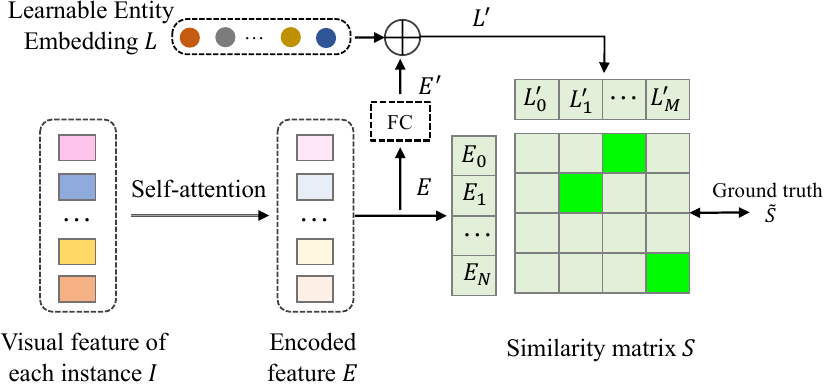}
	\end{center}
	\caption{The structure of the proposed contrast-guided feature adjustment module (CFAM).}
	\label{fig:ICLM}
\end{figure}

Contrastive learning is first proposed to bridge the tasks of OCR and information extraction. The major difficulty is how to design proper representations on behalf of the two tasks. Therefore, we propose CFAM, the key component of our framework, as shown in Figure~\ref{fig:ICLM}. Given visual features of each instance $I \in \mathbb{R}^{N \times C}$, the encoded features $E\in \mathbb{R}^{N \times C}$ are obtained via context modeling between $I$, where $N$ and $C$ indicate the number of instances and the channel of features, respectively.
Then $E$ is used to guide the generation of entity features $L'\in \mathbb{R}^{M \times C}$, where $M$ indicates the number of entity categories. Next, we take $E$ and $L'$ as inputs and calculate a similarity matrix $S\in \mathbb{R}^{N \times M}$ between them. To supervise the generation of $S$, we use the ground truth $\tilde{S}$ of correspondences between instances and entity categories, which is obtained from the ground truth of the entity. Finally, the outputs of CFAM are $S$ and $E$.

\paragraph{Generation of encoded features.}
The visual features of each instance $I$ usually are discrete among instances. Thus, we use three Self-Attention ($SA$) layers to encode $I\in \mathbb{R}^{N \times C}$ and generate the encoded features $E\in \mathbb{R}^{N \times C}$, which can effectively establish connections among instances and transfer rich visual features $I$ from the text detection to the information extraction. The $SA$ consists of three inputs, including queries ($Q$), keys ($K$), and values ($V$), defined as follows:
\begin{equation}
SA(Q,K,V) = softmax(\frac{Q\cdot{K^T}}{\sqrt{C/m}})\cdot{V},
\label{eq:SA}
\end{equation}
where $Q$, $K$, and $V$ are obtained from the same input $I$ (e.g., $Q$=$IW_Q$). Particularly, we use the multi-head self-attention ($MSA$) to construct the complex feature relations, $MSA = [SA_1;SA_2;..;SA_m]W_O$, where $W_O$ is a projection matrix and $m$ is the number of attention heads, set as 8.

\paragraph{Generation of entity features.}
It is crucial to design the entity features $L'\in \mathbb{R}^{M \times C}$, which represents the information extraction task. The natural way is directly adopting the learnable entity embeddings $L\in \mathbb{R}^{M \times C}$ as entity features. However, the entity features are almost the same among all samples for the same initialization of learnable entity embeddings and cannot fully use the information from OCR. Additionally, simply using encoded features $E$ as entity features causes lower generalization for information extraction. Thus, based on this consideration, we adopt $E$ and $L$ together to generate the entity features $L'$. Given the encoded features $E$, which are first transposed as $E^T$ 
, then the $E^T$ are fed into $FC$ layer, which results in $E'\in \mathbb{R}^{C \times M}$: 
\begin{equation}
E' = FC(E^T)
\label{eq:Embedding}
\end{equation}
Next, we transform the $E'$, then use $E'$ to guide the generation of the entity features $L'\in \mathbb{R}^{M \times C}$ by adding the $E'$ to learnable entity embeddings $L\in \mathbb{R}^{M \times C}$, defined as follows:
\begin{equation}
L' = L + E'{}^T
\label{eq:Embedding}
\end{equation}

\paragraph{Generation of similarity matrix.}
The correspondences between encoded features $E$ and entity features $L'$ are significant for establishing the relations between the tasks of OCR and information extraction. Hence, we take $E$ and $L'$ jointly as inputs to construct the relations between them via calculating the product of the two matrices, which results in a similarity matrix $S\in \mathbb{R}^{N \times M}$ reflecting the relations between encoded features $E$ and entity features $L'$:
\begin{equation}
S = E \cdot L'^T
\label{eq:matmul}
\end{equation}

Moreover, we use the ground truth $\tilde{S}$ that is obtained from the annotated labels of the entity to supervise the generation of $S$.

\subsection{Information Extraction Module}
Information extraction requires both visual and textual understanding to automatically assign visual elements to various entities.
Therefore, the recognition features $T$, encoded features $E$, and similarity matrix $S$, indicating the visual, textual and correlative information are used for information extraction. Following previous representative methods~\cite{wang2021towards,zhang2020trie}, we adopt LSTM-CRF\cite{lample2016neural} as an information extraction module to predict entity categories for all characters at the character level and output the final structure results.

\subsection{Loss Function}
In the training phase, our proposed framework can be trained in an end-to-end manner with the weighted sum of the losses generated from four parts of text detection, recognition, information extraction, and contrastive learning, which are defined as follows:
\begin{equation}
L = \alpha L_{det} + \beta L_{rec} + \gamma L_{ie} + \lambda L_{c}, 
\label{eq:matmul}
\end{equation}
where $\alpha$, $\beta$, $\gamma$, and $\lambda$ are hyper-parameters that control the tradeoff between losses. 

$L_{det}$ is the same loss as Mask R-CNN~\cite{he2017mask} for text detection. $L_{rec}$ is the cross entropy loss for character recognition as follows:
\begin{equation}
L_{rec} = CE(R, \tilde{R}), 
\label{eq:matmul}
\end{equation}
where $CE$ indicates the cross entropy loss, $\tilde{R}$ denotes the ground truth of each character.

$L_{ie}$ is also the cross entropy loss for entity classification as follows:
\begin{equation}
L_{ie} = CE(A, \tilde{A}), 
\label{eq:matmul}
\end{equation}
where $CE$ indicates the cross entropy loss, $\tilde{A}$ denotes the ground truth of entity category.

$L_{c}$ indicates the loss for contrastive learning. Due to the imbalanced distribution of the number of entities, we adopt the focal loss~\cite{lin2017focal} for $L_{c}$ as follows:
\begin{equation}
L_c = FL(S, \tilde{S}), 
\label{eq:matmul}
\end{equation}
where $FL$ indicates the focal loss, $\tilde{S}$ denotes the actual entity category of each character.

\begin{table*}[t]
\setlength{\tabcolsep}{1.5mm}
\footnotesize
\centering
\caption{Performance comparison of the SOTA algorithms on the SROIE and POIE datasets. For text spotting, the results contain two parts (the left and right parts indicate the results of text detection and recognition, respectively). The $\Delta$ means the drop of results from SROIE to POIE dataset. * indicates our reproduced results. Note that all results are exhibited in F1-Score.}
\label{tab:end-to-end_settings}
\vspace{5pt}
\begin{tabular}{ ccccc }
\toprule
Setting&Method& SROIE& POIE & $\Delta$ \\
\midrule
Text Spotting&Mask Textspotter~\cite{liao2020mask}*&97.38/91.23&97.68/91.03&0.30/-0.20\\
\midrule
{\multirow{2}{*}{Pure IE}}&VIES~\cite{wang2021towards} &96.12&84.08*&-12.04\\
&PICK~\cite{yu2021pick}&96.12&83.23*&-12.89\\
&TRIE~\cite{zhang2020trie} &96.18&82.45*&-13.73\\
&LayoutLMv2~\cite{xu2021layoutlmv2}&96.25&84.18&-12.07\\
\midrule
\multirow{2}{*}{End-to-End IE}&TRIE~\cite{zhang2020trie} &82.06&76.37*&-5.69\\
&VIES~\cite{wang2021towards}* &83.44&77.19&-6.25\\
\bottomrule
\end{tabular}
\end{table*}

\section{Experiments}
\subsection{Implementation Details}
Our proposed method is implemented in Pytorch. We use four TITAN Xp with 12GB RAM to train our model with batch size four and the Adadelta optimizer. The learning rate starts from 2e-4 and decays to 2e-6 following the linear decay schedule. For the SROIE dataset, the total training epoch is set to 600 for a fair comparison. The total training epoch for the POIE is set to 200 for a fair comparison. We adopt Mask R-CNN~\cite{he2017mask} as our text detector with ResNet-50~\cite{he2016deep} followed by FPN~\cite{lin2017feature}. The hyper-parameters $\alpha$ and $\beta$ are set to 1.0 while $\gamma$ and $\lambda$ are set to 10.0. We choose the best results of all methods in total epoch as the final results. Besides, we use bounding boxes and transcripts for all the text in the images.

\subsection{Analysis of Our Proposed Dataset}
In this part, to verify the practical utility of our POIE dataset, we make comprehensive comparisons between the POIE dataset and the SROIE~\cite{huang2019icdar2019} dataset. Although SROIE is the most widely used English dataset in the field of VIE, there exists a large number of errors in the annotations of SROIE. Additionally, the SROIE mainly focuses on document images with a single scene layout and few entities, which cannot fully reflect the challenges of information extraction in the wild. The comparisons are from three aspects: 1) We first explore the performance of popular text spotting method~\cite{liao2020mask} on these two datasets. The aim of this setting is to verify the differences between the OCR task and VIE task. 2) Then, we evaluate the performance of a few typical methods for the pure IE task (i.e., directly using the ground truth of bounding boxes and texts as the inputs for information extraction). This setting can effectively reveal the difficulty of variable layouts and entities for information extraction. 3) Finally, to further evaluate the challenges of our dataset, we evaluate some methods in the end-to-end setting (i.e., using OCR results from text spotting as inputs for information extraction.) The details comparisons of these three settings are listed as follows:
\paragraph{Text spotting setting.}
As shown in the first part of Table~\ref{tab:end-to-end_settings}, we observe that the performances of the pioneering text spotting algorithm Mask Textspotter~\cite{liao2020mask} are slightly different on POIE and SROIE, which demonstrates that owing to the rapid development of text spotting, document and scene texts can be effectively detected and recognized by the advanced algorithms.

\paragraph{Pure IE setting.}
As listed in the second part of Table~\ref{tab:end-to-end_settings}, we find that the performance of representative methods~\cite{zhang2020trie,wang2021towards} have a significant drop from SROIE to our dataset, e.g., the performance of VIES and TRIE from 96.12 to 82.91 and from 96.18 to 81.79 respectively. The reason is that the variable layouts and entities are much more difficult for information extraction, demonstrating that our dataset is more practical for promoting advanced VIE algorithms.

\paragraph{End-to-end IE setting.}
As shown in the third part of Table~\ref{tab:end-to-end_settings}, the performance of the TRIE~\cite{zhang2020trie} and VIE~\cite{wang2021towards} has distinct differences from SROIE to ours, e.g., the performance of TRIE and VIES from 82.06 to 76.37 and from 83.44 to 77.19, respectively, which further proves our dataset is also more practical for end-to-end methods.

\begin{figure}[t]
	\begin{center}
		\includegraphics[width=0.96\linewidth]{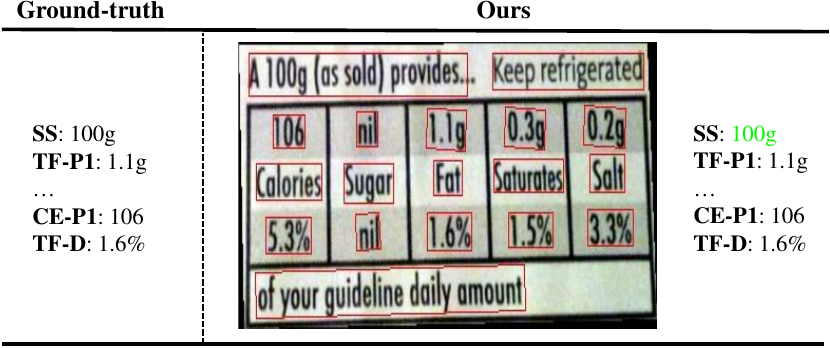}
	\end{center}
	\vspace{-15pt}
	\caption{Qualitative analysis of the proposed POIE dataset. \textcolor{red}{Red} boxes indicate the predicted detection results. Texts show the results of information extraction. Note that \textcolor{green}{green} texts indicate the false negative results.}
	\label{fig:difficult_images}
\end{figure}

\paragraph{Qualitative analysis.}
In this part, we show the qualitative analysis for the difficulty of our POIE dataset. To extract the values of entities from images, the method requires adequate semantic understanding, which is very challenging for algorithms. A typical challenge of our POIE dataset can be seen in Figure~\ref{fig:difficult_images}. We find that without any clues for the ``serving size" (an entity in our dataset), the algorithm cannot accurately predict the value of ``serving size".

\begin{table}[t]
\setlength{\tabcolsep}{3.0mm}
\footnotesize
\centering
\caption{Performance comparison of the state-of-the-art algorithms on SROIE and POIE datasets. * indicates our reproduced results. Backbone of all algorithms is ResNet50. Note that all methods are evaluated in F1-Score without post-processing.}
\label{tab:compared_sota}
\vspace{5pt}
\begin{tabular}{ ccc }
\toprule
Method& SROIE& POIE \\
\midrule
GRAPHIE~\cite{qian2019graphie}&76.51&-\\
Chargrid~\cite{katti2018chargrid}&78.24&-\\
GCN~\cite{liu2019graph}&80.76&-\\
TRIE~\cite{zhang2020trie}&82.06&-\\
\midrule
TRIE~\cite{zhang2020trie}*&82.82&76.37\\
VIES~\cite{wang2021towards}*&83.44 &77.19\\
\textbf{Ours}&\textbf{85.87}&\textbf{79.54}\\
\bottomrule
\end{tabular}
\end{table}

\subsection{The Comparisons of Our Method and SOTA}
In this part, we compare our method with previous state-of-the-art (SOTA) methods~\cite{qian2019graphie,katti2018chargrid,liu2019graph,zhang2020trie,wang2021towards} on POIE and SROIE datasets. Our method is in an end-to-end manner, and the core of our approach is to narrow the semantic gap between the tasks of OCR and information extraction. Consequently, we mainly compare our proposed method with the end-to-end methods and focus on the performance of IE (i.e., F1-score of information extraction).

\begin{table}[t]
\footnotesize
\setlength{\tabcolsep}{1.5mm}
\centering
\caption{Ablation study of CFAM on the proposed POIE dataset. SCH indicates a simple classification head.}
\label{tab:Ablation study of CFAM}
\begin{tabular}{ ccccc }
\toprule
\multirow{2}{*}{SCH}&\multirow{2}{*}{CFAM}&\multicolumn{3}{c}{POIE dataset}\\
 \cmidrule{3-5}
 &&F1-Score (DET)& F1-Score (REC) & F1-Score (IE)\\
\midrule
-&-&97.50&89.40&75.05\\
\checkmark&-&97.10&89.01&77.59\\
-&\checkmark&\textbf{97.89}&\textbf{91.68}&\textbf{79.54}\\
\bottomrule
\end{tabular}
\end{table}

\begin{figure*}[t]
	\begin{center}
		\includegraphics[width=0.96\linewidth]{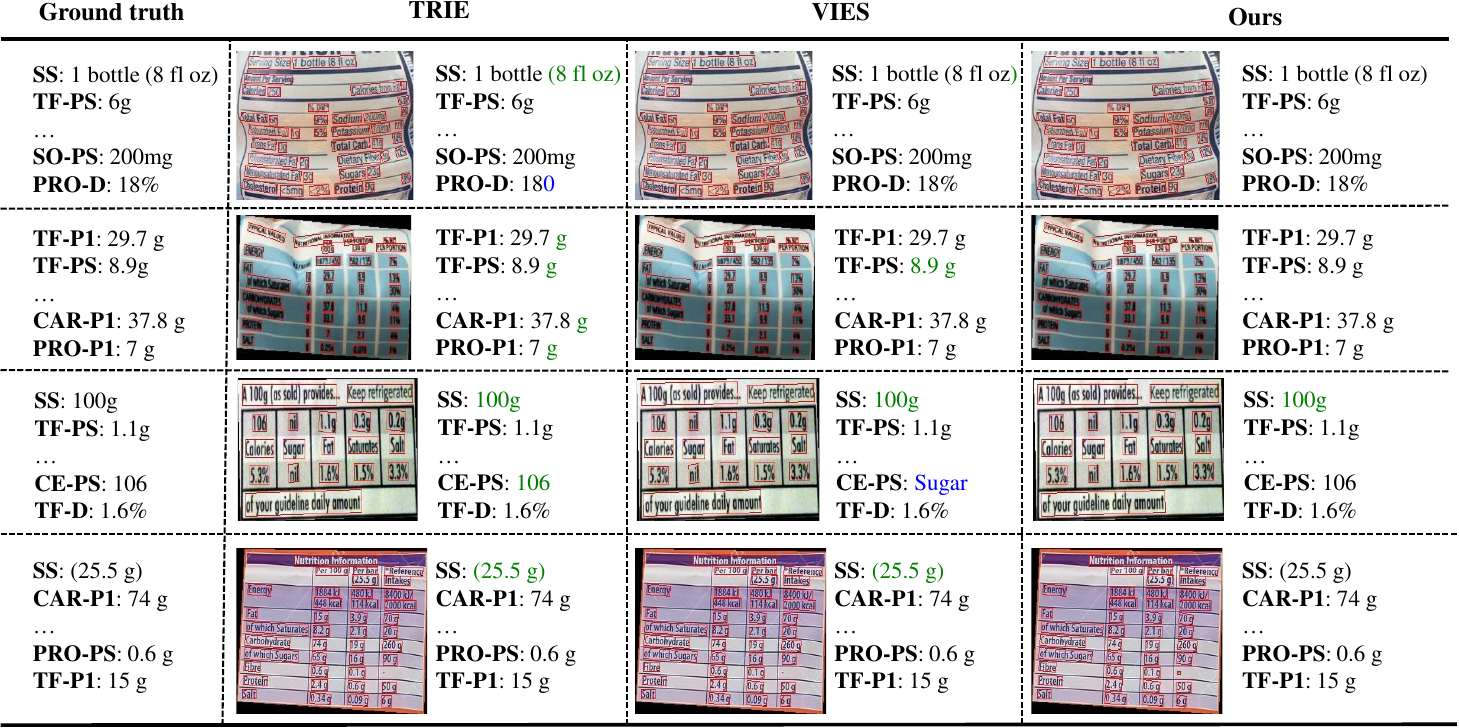}
	\end{center}
\caption{Some visualization results of our method and others on the proposed POIE dataset. There are results of TRIE, VIES, and ours, from left to right, respectively. \textcolor{red}{Red} boxes indicate the predicted detection results. Texts show the information extraction results. Note that \textcolor{green}{green} and \textcolor{blue}{blue} indicate the false negative and false positive results, respectively.}
	\label{fig:visual_result}
\end{figure*}

As shown in Table~\ref{tab:compared_sota}, our method achieves SOTA results on POIE and SROIE datasets, and outperforms other methods by an obvious margin. For the POIE dataset, our method exhibits superior performance. For the SROIE dataset, most previous methods do not use post-progress, thus we mainly focus on the results produced without post-progress. Additionally, our method outperforms other methods by obvious gains. Moreover, Figure~\ref{fig:visual_result} shows some qualitative results of ours and other methods. We can observe that our method is more precisely to predict the values of entities.

\subsection{Ablation Studies}
The ablation studies are carried out on our POIE dataset. For a complete comparison, we show the experimental results of DET, REC, and IE to evaluate the effectiveness of our proposed modules. Specifically, as described below:

\paragraph{Effectiveness of CFAM.} To verify the effectiveness of CFAM, we conduct experiments in the settings: classification and CFAM. Classification indicates directly classifying visual features to the corresponding entity. As listed in Table ~\ref{tab:Ablation study of CFAM}, we observe that classification can promote the performance of information extraction but have a negative impact on text detection and recognition tasks. We argue that it is because our method is an end-to-end method, where the OCR and information extraction tasks are joint optimizations. Moreover, compared with classification, we find that with CFAM, our method achieves superior results (97.89\% in DET, 91.68\% in REC, 79.54\% in IE) for the reason that CFAM serves as an auxiliary module to bridge the relations between the tasks of OCR and information extraction. Additionally, to verify the necessity of using supervision for contrastive learning in CFAM, we conduct experiments in the setting: with $L_{c}$ and without $L_{c}$. As shown in Table~\ref{tab:Ablation study of contrastive loss}, we found that the contrastive loss $L_{c}$ can promote the performance.

\begin{table}[t]
\footnotesize
\setlength{\tabcolsep}{1.5mm}
\centering
\caption{The effectiveness of $L_{c}$ in CFAM.}
\label{tab:Ablation study of contrastive loss}
\begin{tabular}{ cccc }
\toprule
\multirow{2}{*}{$L_{c}$}&\multicolumn{3}{c}{POIE dataset}\\
 \cmidrule{2-4}
 &F1-Score (DET)& F1-Score (REC) & F1-Score (IE)\\
\midrule
-&97.48&91.54&78.69\\
\checkmark&\textbf{97.89}&\textbf{91.68}&\textbf{79.54}\\
\bottomrule
\end{tabular}
\end{table}

\begin{table}[t]
\footnotesize
\setlength{\tabcolsep}{0.5mm}
\centering
\caption{Analysis of generated entity features. Note that all results are shown in F1-Score.}
\label{tab:Ablation study of the generation of entity feature}
\begin{tabular}{ ccccc }
\toprule
\multirow{2}{*}{Learnable entity embeddings}&\multirow{2}{*}{Encoded features}&\multicolumn{3}{c}{POIE dataset}\\
 \cmidrule{3-5}
 &&DET& REC & IE\\
\midrule
\checkmark&-&97.48&90.51&78.86\\
-&\checkmark&97.81&91.48&76.25\\
\checkmark&\checkmark&\textbf{97.89}&\textbf{91.68}&\textbf{79.54}\\
\bottomrule
\end{tabular}
\end{table}

\begin{table}[t]
\setlength{\tabcolsep}{1.5mm}
\footnotesize
\centering
\vspace{5pt}
\caption{The effectiveness of CFAM in other methods on the SROIE and POIE datasets. We reproduce the TRIE and VIES. Note that all results are exhibited in F1-Score.}
\label{tab:CFAM in other methods}
\begin{tabular}{ cccc }
\toprule
Method& SROIE& POIE \\
\midrule
TRIE~\cite{zhang2020trie} &82.82&76.37\\
TRIE~\cite{zhang2020trie} + CFAM (\textbf{ours}) &\textbf{84.25}&\textbf{79.17} \\
\midrule
VIES~\cite{wang2021towards} &83.44&77.19\\
VIES~\cite{wang2021towards} + CFAM (\textbf{ours}) &\textbf{84.32}&\textbf{79.11} \\
\bottomrule
\end{tabular}
\end{table}

\paragraph{Analysis of generated entity features.} In our proposed CFAM module, it is important to design the entity features representing the information extraction task. There are three strategies for generating the entity features (i.e., directly using learnable entity embeddings, adopting encoded features, and combining the learnable entity embeddings and encoded features). The performance of these strategies is shown in Table~\ref{tab:Ablation study of the generation of entity feature}. We observe that using learnable entity embeddings and encoded features together achieves the best results. However, either using learnable entity embeddings or encoded features will get lower results. We argue that the reasons are two-fold: 1) without encoded features, the generated entity features are almost the same among all entity features and cannot fully use the information from instance. 2) without learnable entity embeddings, the generated entity features completely rely on the encoded feature and cannot fully represent the feature of information extraction.

\paragraph{Generalization of CFAM.}
To further demonstrate the generalization of our proposed CFAM, we evaluate CFAM by applying it to other SOTA methods~\cite{zhang2020trie,wang2021towards}. As listed in Table~\ref{tab:CFAM in other methods}, the proposed CFAM improves the performance of TRIE and VIES by a distinguish margin even though they are strong methods. For our POIE dataset, CFAM improves the SOTA method TRIE and VIES by 2.45\% and 1.92\%, respectively. For the SROIE dataset, CFAM enhances the performance of TRIE and VIES by an obvious margin. The results demonstrate the generalization of our proposed CFAM, which can be effectively applied to other methods.

\subsection{Limitation}
Our method has achieved superior results on SROIE and ours. However, we find that the challenges, e.g., different forms for the same entity (Figure~\ref{fig:datasets} (b)), have not been solved well and still influence the performance of information extraction. In the future, we will further explore the commonality between different forms of the same entity and design a new method to effectively solve this problem.

\section{Conclusion}
In this paper, we have proposed a large-scale English dataset (called POIE) consisting of camera images, which can reflect the challenges of the real world. We have designed a novel end-to-end framework with a plug-and-play CFAM for VIE tasks, which adopts contrastive learning and properly designs the representation of VIE tasks for contrastive learning. The experimental results prove that our dataset is more practical for promoting advanced VIE algorithms. Additionally, the experiments demonstrate that our proposed framework consistently achieves the obvious performance gains on SROIE and ours. In the future, we will consider to extend the dataset with more images, entities, and diverse natural scene disturbances. We hope our proposed dataset and framework can promote further investigation in VIE.

\textbf{Acknowledgements.} This work was supported by the National Science Fund for Distinguished Young Scholars of China (Grant No.62225603).

\bibliographystyle{splncs04}
\bibliography{reference}





\end{document}